\documentclass{article} 

\usepackage{arxiv}
\usepackage{amsfonts}
\usepackage{subfigure} 
\usepackage{multirow}
\usepackage{natbib}
\usepackage{booktabs}
\usepackage{siunitx}
\usepackage{comment}
\usepackage{algorithmic}
\usepackage{algorithm}
\usepackage{amsmath}
\usepackage{bbm}
\usepackage{dsfont}
\usepackage{mathtools}
\usepackage{todonotes}
\usepackage[T1]{fontenc}
\usepackage{amsthm}
\usepackage{thmtools}
\usepackage{comment}
\usepackage{adjustbox}
\usepackage{authblk}
\usepackage[utf8]{inputenc}

\newtheorem{mythm}{Theorem}

\newtheorem{myprops}{Proposition}
\newtheorem{mydef}{Definition}

\newtheorem{myex}{Example}
\newtheorem{theorem}{Theorem}

\newtheorem{remark}{Remark}

\DeclareMathOperator{\E}{\mathbb{E}}
\DeclareMathOperator{\R}{\mathbb{R}}

\newcommand{\ex}{\phi}
\newcommand{\f}{v}

\newcommand{\x}{\mathbf{x}}
\newcommand{\p}{f}
\newcommand{\prob}{p}

\begin{document}

\title{Threading the Needle of On and Off-Manifold Value Functions for Shapley Explanations}
\date{}
\author[1]{Chih-Kuan Yeh}
\author[2]{Kuan-Yun Lee}
\author[3]{Frederick Liu}
\author[1]{Pradeep Ravikumar}
\affil[1]{Machine Learning Department, Carnegie Mellon University}
\affil[2]{University of California, Berkeley}
\affil[3]{Google Inc.}
\maketitle

\begin{abstract}
\vspace{0mm}
    A popular explainable AI (XAI) approach to quantify feature importance of a given model is via Shapley values. These Shapley values arose in cooperative games, and hence a critical ingredient to compute these in an XAI context is a so-called value function, that computes the ``value'' of a subset of features, and which connects machine learning models to cooperative games.  There are many possible choices for such value functions, which broadly fall into two categories: on-manifold and off-manifold value functions, which take an observational and an interventional viewpoint respectively. Both these classes however have their respective flaws, where on-manifold value functions violate key axiomatic properties and are computationally expensive, while off-manifold value functions pay less heed to the data manifold and evaluate the model on regions for which it wasn't trained. Thus, there is no consensus on which class of value functions to use. In this paper, we show that in addition to these existing issues, both classes of value functions are prone to adversarial manipulations on low density regions. We formalize the desiderata of value functions that respect both the model and the data manifold in a set of axioms and are robust to perturbation on off-manifold regions, and show that there exists a unique value function that satisfies these axioms, which we term the Joint Baseline value function, and the resulting Shapley value the Joint Baseline Shapley (JBshap), and validate the effectiveness of JBshap in experiments.
\end{abstract}
\section{Introduction}
\label{sec:intro}
\vspace{-3mm}

Shapley values \citep{shapley_1988} were originally proposed to measure the contributions of players in the context of cooperative games. They have since gained significant traction in the area of \emph{explainable machine learning}, where they are used to measuring contributions of different features \citep{gromping2007estimators, lindeman1980introduction, owen2014sobol, owen2017shapley, datta2016algorithmic, lundberg2017unified}, data points \citep{ghorbani2019data, jia2019towards}, neurons \citep{ghorbani2020neuron}, and concepts \citep{yeh2020completeness}, towards the output of a learned model. 

Despite their popularity, using Shapley values in an explainable machine learning context is not straightforward. In a cooperative game setting, the game is characterized by a set value function that takes as input a subset (also called a coalition) of players, and outputs the \emph{value} or utility of this subset. Given this set value function, Shapley values then provide real-valued player-wise contributions. In a machine learning context, however, there is no well-defined set value function that takes as input a subset of features and outputs the value of this subset. Instead, we have the model $\p$, data distribution $\prob$, and the test data point $\x$. A critical ingredient in order to derive Shapley values is to define a ``value function'' $\f_{\x,\p,\prob}(S)$ which takes a set of features $S$ as input, along with the model $\p$, the test data point $\x$, and data distribution $\prob$.
The value function describes how $\p(\x)$ would change if only a set of features in $\x$ participates in an cooperative setting.

However, there is no consensus on how to specify the value function. Existing value functions can be roughly classified into two classes: so-called \emph{off-manifold} value functions measure the utility of coalitions of features by perturbing features outside the coalition (and hence taking the inputs outside the data manifold), while so-called \emph{on-manifold} value functions measure the utility of coalitions of features by marginalizing out features outside the coalition while respecting the data distribution/manifold.
Both approaches come with their respective caveats \citep{kumar2020problems}. Off-manifold value functions by construction evaluate the model on data that is not from the training data distribution; and hence are sensitive to how the model handles off-manifold test data~\citep{hooker2019please, slack2020fooling}, and where machine learning models are typically unstable~\citep{frye2020shapley}. Moreover, the resulting explanations are very susceptible to manipulation: one can manipulate the model function only on off-manifold regions while preserving its behavior on-manifold, just to get the desired explanation (e.g. to show that the model is unbiased, while maintaining biased behavior on-manifold)~
\citep{slack2020fooling}. In other words, the off-manifold value functions respect the model function, but not the data distribution. This is problematic since these value functions evaluate the model outside the data manifold, where most ML training methods do not come with any guarantees on model performance. 

On-manifold value functions on the other hand are in general computationally expensive, particularly if they involve conditional expectations due to marginalizing over out-of-coalition features while conditioning on the coalition features. Approximating these conditional expectations via empirical distributions could result in idiosyncratic issues, such as all features having the same attribution~\citep{sundararajan2019many}. \citet{lundberg2017unified, aas2019explaining, vstrumbelj2014explaining} compute the conditional value function via distributional assumptions, such as independence and Gaussian mixtures, which may not be flexible enough for complex data. \citet{lundberg2018consistent} attempt to calculate the conditional expectation value function for tree ensembles, but which \citet{sundararajan2019many} criticize as having unclear assumptions on the features. \citet{frye2020shapley} propose approximating these conditional expectations via a supervised and an unsupervised approach. As we show however, the supervised approach converges to the empirical conditional value function, while the unsupervised approach generates samples whose coalition features need not be set to the conditioned values. Another principled approach of calculating the conditional expectation based value function is by using importance sampling on the joint distribution, but this is computationally expensive.  Further, we show that conditional expectation based value functions, even when calculated accurately, will also be prone to adversarial manipulation off the data manifold. Another caveat with on-manifold value functions is that they result in Shapley values that violates natural axioms~\citep{sundararajan2019many}. Moreover, they might respect the data distribution at the price of fidelity to the model function: a feature with no intervention effect can still receive non-zero feature importance, which was seen as a positive aspect by \citet{adler2018auditing}, but a negative aspect by \citet{merrick2020explanation, sundararajan2019many}.

So both on-manifold and off-manifold value functions, and their resulting Shapley value explanations, have their competing pros and cons, and \citet{chen2020true} among others have suggested to ``pick one's poison'' when faced with a specific application. \citet{kumar2020problems} have even suggested that these two competing approaches present a serious concern with using Shapley values itself. But what if we can design value functions that respect both the model and the data manifold, and are not susceptible to adversarial manipulations in low density data regions? Surprisingly, we show that we indeed can. Towards this, we first propose a set of axioms that aim to formalize the desiderata in the question above, and show that there exists a unique value function that satisfies these axioms, which we call the Joint Baseline value function, satisfies these axioms. While the Shapley value by Joint Baseline value function does not satisfy the original set of axioms proposed in \citet{sundararajan2019many}, it does satisfy a new set of axioms which takes both data density and function model into account. Moreover, we show that this unique value function is robust to low density adversarial manipulations. We also show that the resulting Shapley value, Joint Baseline Shapley (JBshap), can be computed efficiently and scaled to high dimensional data such as images.



\vspace{-2mm}

\section{Related Work}

\vspace{0mm}

Our work focuses on feature-based explanations, where we assign real-valued attributions to features with respect to their contributions towards the prediction of the model at a given test input. Feature-based explanations include perturbation-based attributions \citep{zeiler2014visualizing, ribeiro2016should}, and gradient-based explanations \citep{simonyan2013deep, selvaraju2016grad, shrikumar2017learning, sundararajan2017axiomatic}. Shapley value belongs to perturbation-based attributions~\citep{ancona2017unified, yeh2019on}. Other forms of explanations include sample-based explanation methods which explains the decision of the model via attributions to training samples \citep{koh2017understanding, DBLP:conf/nips/YehKYR18}, and concept-based explanation methods which assign attributions to high-level human concepts \citep{kim2018interpretability, ghorbani2019towards, bau2017network}. 

Our work specifies a new set of axioms on the set value functions to be used with the Shapley value, which is novel to the best of our knowledge. To position our development, we briefly discuss the axiomatic treatment of Shapley values and extensions, as an approach for real-valued player-wise attributions. Shapley values originated in a cooperative game context, where we are given a set function that specifies the utility of any subset of players. Here, Shapley values were shown to uniquely satisfy a set of natural axioms~\citep{shapley_1988}; and were discussed in an explainable AI context by \citep{lundberg2017unified}.  In a more general so-called cost-sharing game context, we are given a cost-sharing function that goes from (non-negative) real-valued demands associated with players to the overall cost, and a number of extensions of Shapley values such as Aumann-Shapley and Shapley-Shubik have been developed to uniquely satisfy their respective sets of axioms~\citep{friedman1999three, aumann2015values}; and were discussed in an explainable AI context \citep{sundararajan2019many}. Our axiomatic developments on the other hand are complementary to these classical developments: we focus specifically on \emph{how to reduce the explainable AI problem to a cooperative game setting}, upon which we could then use Shapley values that were actually developed in the context of such games.

\vspace{-4mm}
\section{Different Value Functions for Shapley Value Explanations} \label{sec:values}
\vspace{-2mm}
\subsection{Problem Definition}
\vspace{-2mm}
Given a machine learning model $\p(\cdot): \R^d \rightarrow \R$ trained on $d$-dimensional data, 
data distribution $\prob$, and an {\it explicand} (i.e. test input) $\x \in \mathbb{R}^d$, we aim to attribute the model output $f(\x)$ to the individual features. We use $\ex_{\p,\prob,i}(\x)$ to denote this {\it attribution} to the $i$-th feature. In certain cases, there is also a given baseline value $\x' \in \mathbb{R}^d$, which is often set to $0$ or the data mean. When applying Shapley values (which originate in cooperative game theory) to this attribution problem, a critical first step is to capture the key elements of $\p,\prob,\x$ into a set value function $\f_{\x,\p,\prob}(S): 2^d \rightarrow \mathbb{R}$, where $S \subseteq [d]$ is a subset of features. In this context, subsets of features can be thought of as subsets of players; in the sequel, we will often refer to the $i$-th feature as the $i$-th {\it player} to emphasize the connection to game theory at play with Shapley explanations. Given such a set value function $v(\cdot)$, Shapley value attributions to the $i$-th player can be computed as $\phi_{i}(v) = \!\!\! \sum_{S \subseteq [d]/ i} \!\!\!\! \frac{s! (n-s-1)!}{n!}[v(S \cup i) - v(S)]$.
 In other words, the Shapley value of player $i$ is the weighted average of its marginal contribution $v(S\cup i) - v(S)$ averaged over all possible subsets $S \subseteq [d]$ that do not include player $i$. The Shapley value has been shown to be the unique solution satisfying a set of reasonable axioms; see \citep{shapley_1988} for a further discussion and history. We will provide a more detailed discussion of such axioms in Sec. \ref{sec:new_axiom} and Appendix~\ref{sec:translative}.
 
In the explainable AI context, using the set value function $\f_{\x,\p,\prob}(\cdot)$, we can then compute the Shapley importance values $\ex_{\p,\prob,i}(\x) = \ex_i(\f_{\x,\p,\prob}(\cdot))$, where $\ex_i$ is the Shapley value attribution to the $i$-th player as detailed above. The critical question however is: how to specify the set value function $\f_{\x,\p,\prob}(\cdot)$?


\vspace{-2mm}
\subsection{Notation for Existing On-Manifold and Off-Manifold Value Functions}
\vspace{-2mm}

Different choices of the set value function $v$ would lead to different Shapley values $\phi_i(v)$, and indeed the proper selection of $v$ has been of great interest within the explainable AI community. We first set up some notation. We let $x_S$ denote the sub-vector of $\x$ corresponding to the feature set $S \subseteq [d]$, and $x_{\bar{S}}$ as the sub-vector corresponding to the complementary subset $\bar{S} \subseteq [d]$. Given a {\it baseline} $\x' \in \mathbb{R}^d$, we will be using the notation $f(x_S; x'_{\bar{S}})$ to refer to the function $f$ with input sub-vectors $x_S$ and $x'_{\bar{S}}$ corresponding to features in $S$ and $\bar{S}$ respectively.
\vspace{-2mm}
\subsubsection{On-Manifold Value Functions}
\vspace{-2mm}
In this paper, we consider the prototypical on-manifold value function of Conditional Expectation, which in turn results in the Conditional Expectation Shapley value ({\it CES})~\citep{lundberg2017unified, sundararajan2019many}. We will be overloading notation and term the \emph{value function} itself as CES.

\vspace{0mm}
\paragraph{\bf CES:} $\f^C_{\x, \p, \prob}(S) = \E_{\x'_{\Bar{S}} \sim \prob(\x'_{\Bar{S}}| \x_S)} [\p(\x_S; \x'_{\Bar{S}})] = \int_{\x'_{\Bar{S}} }\p(\x_S; \x'_{\Bar{S}}) p(\x'_{\Bar{S}}| \x_S) d \x'_{\Bar{S}}$. In other words, CES takes the expectation of $f(\x_S;\x_{\bar{S}}')$ over the conditional density of $\x_{\bar{S}}'$ given $\x_S$ (with respect to the joint density $p(\cdot)$). 
\vspace{-2mm}
\subsubsection{Off-Manifold Value Functions}
\vspace{-2mm}
As discussed earlier, off-manifold value functions quantify the contribution of a coalition of features by perturbing or intervening on the off-coalition features.  In this paper, we consider the prototypical
off-manifold value functions of Baseline Shapley ({\it Bshap})~\citep{sun2011axiomatic, lundberg2017unified}) and its randomized variant.

\vspace{0mm}
\paragraph {\bf Bshap:} $\f^B_{\x,f,p}(S):= \p(\x_S; \x'_{\Bar{S}} )$. Bshap directly takes the value of $f$ corresponding to $\x$, and ignores the data density model $p$. 
\vspace{0mm}
\paragraph {\bf RBshap:} $\f^{RB}_{\x, \p, \prob}(S) = \E_{\x'_{\Bar{S}}} [\p(\x_S; \x'_{\Bar{S}})]=  \int_{\x'_{\Bar{S}}}\p(\x_S; \x'_{\Bar{S}}) p_{b}(\x'_{\Bar{S}}) d \x'_{\Bar{S}}$, where $p_b$ is the marginal density of $\x_{\bar{S}}$. RBshap is a natural extension of Bshap, where the baseline $\x'$ in Bshap is generalized to a distribution of baselines, which is represented by $p_b(\x')$. 



\vspace{0mm}
\section{Issues of Existing Value Functions}
\label{sec:pitfalls}

\subsection{Off-Manifold Value function: Not ``Respecting'' the Data Manifold}
The main difference between on-manifold value functions and off-manifold value functions is that off-manifold value functions do not take the data manifold (density) into account. The machine learning model $\p$ may not be trained to have meaningful prediction values when the input is $[\x_S; \x'_{\Bar{S}}]$. For instance, for image inputs, $[\x_S; \x'_{\Bar{S}}]$ would represent an image where certain random pixels are replaced by some baseline value (such as setting to black). \citep{frye2020shapley} thus raise the concern that Shapley values calculated by off-manifold value functions could be over-reliant on function values in off-manifold regions, which is not where the model is trained on. Issues could arise when the model behavior is different on the data manifold and off the data manifold. For instance, if the image classifier tends to predict an image with many random black pixels as ``dog'', the Shapley explanation for an image of a dog may be strongly affected by how the model handles such artifacts. This problem is emphasized by \citet{slack2020fooling} as they show that they can train a new model that only differs to the original model in off-manifold regions, and they are able to manipulate the Shapley explanations for off-manifold value functions. We further discuss this issue in Sec. \ref{sec:sensitivity}.

\subsection{On-Manifold Value function: Difficulty to Calculate Conditional Value}

In contrast to the robustness challenges of the previous section, the main difficulty with conditional expectation (CES) based value functions is computational. Many variants have been proposed to simplify the computational burden of calculating the CES value, however, they may come with the cost of its new issues. We discuss some key variants and their respective issues.

\subsubsection{CES-Empirical}  
\vspace{-2mm}
One variant is called CES-Empirical, which uses the empirical probability to calculate the conditioned value. To formally define CES-Empricial, let $\prob^E(\x) = \frac{1}{m} \mathbbm{1}[\x \text{ in dataset}]$ denote the empirical distribution, given $m$ points in the dataset. In this case, $\f^{CE}_{\x, \p, \prob^E}(S) = \E_{\x'_{\Bar{S}} \sim \prob^E(\x'_{\Bar{S}}| \x_S)} [\p(\x_S; \x'_{\Bar{S}})] = \frac{\sum_{i=1}^m \mathbbm{1}[\x_S^i=\x_S]  \p(\x^i)}{\sum_{i=1}^m \mathbbm{1}[\x_S^i=\x_S]}$ will be the average prediction of the points in the dataset where its feature set $S$ is equal to $\x_S$.

Although CES-Empirical can be easy to calculate, the caveats of CES-Empirical have been discussed in a line of recent work; we summarize the key ones below. The main criticism of CES-Empirical is that when an explicand has feature values that are unique in the dataset (which is likely to be the case with a continuous data distribution), each feature will get the same Shapley value importance even if the model is not symmetric in all features~\citep{sundararajan2019many}. We showcase the issues of CES-Empirical with a concrete example in Appendix~\ref{sec:ces_emp_issues}.
\vspace{-2mm}
\subsubsection{CES-Supervised} 
\vspace{-2mm}
Another popular CES version is called CES-Supervised \citep{frye2020shapley}. The method is motivated by the observation that the conditional expectation $\f_{\x,\p,\prob}(S) = \E_{\x'_{\Bar{S}} \sim \prob(\x'_{\Bar{S}}| \x_S)} [\p(\x_S; \x'_{\Bar{S}})]$ can be seen to minimize the MSE loss
$ \text{mse}(v(\x_S)) := \E_{\x'_{\Bar{S}} \sim \prob(\x'_{\Bar{S}}| \x_S)} [\p(\x_S; \x'_{\Bar{S}}) - v(\x_S)]^2.$ Thus, \citet{frye2020shapley} propose to learn the value function by a surrogate model $g$ by minimizing the MSE loss: 
\[\vspace{-2mm} \f^{CS}_{\x, \p, \prob^E}(S) = \arg\min_{g} \E_{\x \sim \prob^E}\E_{S\sim \text{shapley}}[(\p(\x) - g(\x_S))^2], \vspace{-2mm }\]
and the conditioned value $\f^{CS}_{\x, \p, \prob^E}(S)$ can be obtained by the surrogate model $g$ by $g(\x_S)$. \citet{frye2020shapley} propose to use a neural network with masked inputs as the surrogate model.

In order to calculate CES-supervised, \citet{frye2020shapley} follow the standard approach of empirical risk minimization, and optimize the empirical MSE loss:
\vspace{-2mm}
\begin{align}\label{eq:CES_supervised}
    \E_{\x\sim \prob^E(\x)}\E_{S\sim \text{shapley}}[(\p(\x) -\f^{CS}_{\x_S, \p, \prob})^2].
    \vspace{-1mm}
\end{align}

One issue of CES-Supervised is that it requires retraining the model completely, where the sample space is all possible subsets for every input. The training of the model $g$ is even more difficult than the training of the original model $\p$.

Another issue is that the resulting optimal ERM CES-supervised estimator is exactly the CES-Empirical estimate, as we show in the following theorem.

\begin{myprops}
 \label{thm:ces-sup}
The global minimizer to \eqref{eq:CES_supervised}, where $\prob^E(\x) = \frac{1}{m} \mathbbm{1}[\x \text{ in dataset}]$ is exactly equal to $f^{CE}_{\x, \p, \prob^E}(S)$. Thus, $ \f^{CS}_{\x_S, \p, \prob} = f^{CE}_{\x, \p, \prob^E}(S)$ when the empirical distribution is used.
\end{myprops}
\vspace{-1mm}

The proof is in Appendix, which follows from a simple analysis of the stationary conditions of the objectives above. As a consequence of this surprising theorem, all pitfalls of CES-Empirical still carry over to CES-supervised when the empirical distribution is used. Even although \citet{frye2020shapley} use a masked encoder model to learn $\f^{CS}_{\x_S, \p, \prob}$, when the encoder is sufficiently flexible (e.g. large neural models) this still converges to CES-Empirical, and not the true conditioned value. 
One evidence is that CES-supervised is invariant to the behavior of $\p$ off the data points in the dataset. However, the true conditioned expectation should be dependent on the behavior of $\p$ off the data manifold.
We note that when the learning of the surrogate model has strong regularization, CES-SUP may not necessarily converge to CES-Empirical, but the fact that CES-SUP is invariant to the behavior of $\p$ off the data points validates that it does not capture the real conditioned value.
Another issue of CES-Supervised, which is demonstrated in the experiment section, is that the empirical MSE loss can still be low even if the training objective is altered, and thus leaving CES-supervised to be prone to model-based attack.

\vspace{-2mm}
\subsubsection{CES-Sample} \label{sec:cessample}
\vspace{-2mm}
Suppose we have access to the conditioned probability $p(\x'_{\Bar{S}}| \x_S)$, we can use Monte-Carlo sampling to estimate $\f^{CSam}_{\x, \p, \prob}(S) = \int_{\x'_{\Bar{S}} }\p(\x_S; \x'_{\Bar{S}}) p(\x'_{\Bar{S}}| \x_S) d \x'_{\Bar{S}}$. To obtain the conditioned probability, \citet{frye2020shapley} proposed a Masked variational auto-encoder approach, which results in samples that does not respect the conditioned value, as we further discuss in the appendix. An alternative is to sample conditioned probability that satisfies the conditioned constraint is to use importance-sampling, by first sampling uniformly from all $\x'_{\Bar{S}}$, and average the value of $\p(\x_S; \x'_{\Bar{S}}) p(\x'_{\Bar{S}}| \x_S)$ for each sampled point. The main downside of CES-Sample is that it is computationally expensive to calculate $\f^{CSam}_{\x, \p, \prob^E}(S)$, each of which may take around $10$ to $100$ accesses to the model $\p$; in contrast, the computation of most other value functions require single access to the model $\p$. Thus, CES-sample may be $10 \times$ to $100 \times$ more expensive to calculate compared to other value functions.

\vspace{-2mm}
\subsection{Off-Manifold and On-Manifold Value function: Sensitivity to Perturbation In Low-Density Regions}
\label{sec:sensitivity}
\vspace{-2mm}
For off-manifold value functions, it is shown that one can manipulate the Shapley value explanations by perturbing the functions on off-manifold regions \citep{slack2020fooling}. The methodology of \citet{slack2020fooling} is that given an original black-box model $f$ which has explanation $\phi(f)$, one may perturb $f$ to $f'$ by only changing the behavior in low-density regions. In practice, we usually only check the test accuracy and test prediction, and the difference of $f$ and $f'$ may not be detected in several cases. However, the explanation $\phi(f)$ and $\phi(f')$ may differ drastically. This undermines the original explanation $\phi(f)$, as there are now ground truth for how the model $f$ should behave off-manifold. While previous works \citep{frye2020shapley} claim that on-manifold value function is robust to such manipulation, we show that one can also manipulate (the Shapley values corresponding to) on-manifold value functions by perturbing the functions on low density regions (while fixing the model behavior on high density regions).

We first formalize a definition of robustness to off-manifold manipulations, and discuss the sensitivity (robustness) of existing value functions to off-manifold manipulations.

\begin{mydef}
Given any two models $\p_1(\x)$, $\p_2(\x)$ and any probability measure $\prob(\x)$, if $\max_\x |\p_1(\x)-\p_2(\x)| \prob(\x)  \leq \epsilon$ always entails $|\f_{\x,\p_1,\prob}(S) - \f_{\x,\p_2,\prob}(S)| \leq T \epsilon$ for any $S$, we term the value function $\f_{\x,\p,\prob}(S)$ as strong $T\text{-robust}$ to off-manifold perturbations. 
\end{mydef}

The premise $\max_\x |\p_1(\x)-\p_2(\x)| \prob(\x)  \leq \epsilon$ bounds the maximum perturbation on low density regions. For instance, if we require that $\max_\x |\p_1(\x)-\p_2(\x)| \prob(\x)  \leq 0.01$, then in regions with density less than $0.01$, we can perturb the function values by at most $1.0$. However, we show that both Bshap and CES are not strong $T\text{-robust}$.


\begin{myprops} \label{prop:notstrong}
Bshap and CES are not strong $T\text{-robust}$ to off-manifold perturbations for $|T| < \infty$.
\end{myprops}

The key reason that Bshap and CES does not satisfy strong $T\text{-robustness}$ is that the value function $\f_{\x,\p,\prob}(S)$ for Bshap and CES can be determined by the behavior of $\p$ on low density regions. To address this issue, we show that the strong- $T$-robustness can be obtained by taking the data density into account in the calculation of value functions.

We empirically evaluate the sensitivity of Bshap and CES when the function is perturbed in low-density regions in practice in Sec. \ref{sec:experiment}.

\vspace{-2mm}
\section{Axioms for Value Functions \& A New Value Function}
\label{sec:new_axiom}
\vspace{0mm}


In this section, we argue that it is possible to get the best of, and avoiding the pitfalls of, both worlds --- of on-manifold and off-manifold value functions --- by taking both the function and the data distribution into account.

We first propose a set of axioms on set value functions that takes these desiderata into account.
\vspace{-2mm}
\begin{enumerate}
	\item {\bf Linearity (over functions and distributions) (Lin.)}: For any functions $\p, \p_1,\p_2, \prob, \prob_1, \prob_2$, $\alpha_1 \f_{\x,\p_1,\prob}(S) + \alpha_2 \f_{\x,\p_2,\prob}(S) = \f_{\x,\alpha_1 \p_1+\alpha_2 \p_2, \prob}(S)$, for $\alpha_1,\alpha_2 \in \mathbb{R}$ and $\alpha_1 \f_{\x,\p,\alpha_1\prob_1}(S) + \alpha_2 \f_{\x,\p,\prob_1}(S) = \f_{\x,\p, \alpha_1\prob_1 + \alpha_2\prob_2}(S)$ for $\alpha_1, \alpha_2 \ge 0$.

	\item {\bf Symmetry (over functions and distributions) (Sym.)}: for all $i,j \not \subseteq S$,
$\p$, $\prob$, $\x$, $\x'$ all being symmetric in the $i$-th and $j$-th dimensions implies  $\f_{\x,\p,\prob}(i\cup S) = \f_{\x,\p,\prob}(j\cup S)$. That is, if both the function $\p$ and the probability $\prob$ are symmetric with respect to two features, then so should the value function $\f$.

	\item {\bf Dummy Player (over functions and distributions) (Dum.)}: $\p, \prob$ being invariant in the $i$-th dimension implies $\f_{\x,\p,\prob}(S) = \f_{\x,\p,\prob}(i\cup S)$. That is, a completely irrelevant feature in both the function value and data density should also be irrelevant in the value function.
	
	\item {\bf Null Player (over functions and distributions) (Null)}:  if $\p', \prob'$ satisfies $ \p(\x') = \p'(\x'), \prob(\x') = \prob'(\x') $, then $\f_{\p,\prob,\x}(\text{\O}) = \f_{\p',\prob',\x}(\text{\O})$. That is, the value function when no players are in the set function solely depends on the function value and data density of the baseline.
	
	\item {\bf Efficiency (over functions and distributions) (Eff.)}: $ \f_{\p,\prob,\x}[d] - \f_{\p,\prob,\x}[\text{\O}] = \p(x) \prob(\x) - \p(x') \prob(\x')$. The difference between the value function of all players and no players is equal to the total joint density. Note that this axiom is not needed in the uniqueness of JBshap.
	
	\item {\bf Set Relevance (over functions and distributions) (Set.)}: If $ \p_1(\x_S, \bar{x}_{\bar{S}})  = \p_2(\x_S, \bar{x}_{\bar{S}})$ and $ \prob_1(\x_S, \bar{x}_{\bar{S}})  = \prob_2(\x_S, \bar{x}_{\bar{S}})$ for any $\bar{x}$, $\f_{\p_1,\prob_1,\x}(S) = \f_{\p_2,\prob_2,\x}(S)$. The set relevance axiom states that the value function is determined by how $\p, \prob$ behaves when the feature set $S$ is fixed to be equal to $\x_S$.
	
	\item {\bf Strong-T-Robustness (over functions and distributions) (Rob.)}: For any functions $\p_1,\p_2, \prob$, if $\max_\x |\p_1(\x)-\p_2(\x)| \prob(\x)  \leq \epsilon$, then $|\f_{\x,\p_1,\prob}(S) - \f_{\x,\p_2,\prob}(S)| \leq T \epsilon$ for some constant $T$. This axiom has been discussed in Sec. \ref{sec:sensitivity}.
    \vspace{-1.5mm}
\end{enumerate}

\vspace{-1mm}
As discussed earlier, Shapley values originated in a cooperative game context, as the unique \emph{attribution function} that assigns attributions to individual players, given a set value function that specifies utilities of sets/coalitions of players. Our set of axioms above on the other hand are for the set value function itself: on what properties such set-value functions should satisfy. Our linearity axiom states that for a linear combination of models should have a corresponding linear combination of set-value functions. Similarly a convex combination (i.e. mixture) of distributions should have a corresponding convex combination of value functions. 

Note that Shapley values themselves satisfy a linearity axiom, that  states that a linear combination of set-value functions should have a linear combination of corresponding Shapley values. By combining the linearity axioms for set-value functions we specify here, with the linearity axiom of Shapley value feature attributions, it can be seen that we get that a combined linearity axiom: that for linear (convex) combinations $\p$ ($\prob$), the Shapley attributions should in turn be linear (convex) combinations of the corresponding Shapley values. We formalize the complete set of transitive axioms in Appendix Sec. \ref{sec:translative}. 

The set relevance axiom states that the value function is determined by the behavior of $\p$ and $\prob$ when the feature set of $S$ is set to the value of $\x$. The rationale is that the value function identifies the contribution of $\x_S$ in $\p(\x)$ and $\prob(\x)$, and thus should be only relevant to $\p(\bar{\x})$ and $\prob(\bar{\x})$ when $\bar{\x}_S  = \x_S$. In other words, how $\p(\bar{\x}), \prob(\bar{\x})$ behaves when $\bar{\x}_S$ is not equal to $\x_S$ should not affect the value function $\f_{\p,\prob,\x}(S)$. Common value functions also satisfy set relevance axioms, for instance, CES satisfies the (Set.) axiom, and Bshap satisfies the (Set.) axiom with only $\p$ (see (Set-IS.) in Sec. \ref{sec:translative}).

We next show the following uniqueness result:

\begin{theorem} \label{thm:JBshap-uniqueness}
$\f_{\x,\p,\prob}(S)$ satisfies (Lin.), (Sym.), (Dum.),  (Null), (Eff.), (Set.), (Rob.) axioms if and only $\f_{\x,\p,\prob}(S) =  \p(\x_S,\x'_{\Bar{S}})\prob(\x_S,\x'_{\Bar{S}})$.
\end{theorem} 
\vspace{-2mm}

This result specifies the unique value function that satisfies natural axioms which take both the data and model into account. We term this unique value function the {\it Joint Baseline Shapley value} (JBshap):
\begin{mydef} \label{def:JBshap}
(JBshap): \vspace{-2mm}
\begin{align} \label{eq:JBshap}
    \f^J_{\x,\p,\prob}(S) = \p(\x_S; \x'_{\Bar{S}})\prob(\x_S; \x'_{\Bar{S}})
\end{align}

\end{mydef}
\vspace{-1mm}
An interesting fact is that for the uniqueness result, the efficiency axiom is not needed, which can be seen in the proof in Sec.~\ref{sec:proof}.
Suppose the model $f(\x_S; \x'_S)$ converges to the conditional distribution $p(y|\x_S, \x_S')$, then from \eqref{eq:JBshap}, it can be seen that 
$v^{J}_{\x,f,p}(S) \to p(y|\x_S, \x_S') p(\x_S; \x_S') = p(y, \x_S, \x_S').$
In other words, as the model itself converges to the Bayes optimal classifier, JBshap assigns a value to $\x_S$ that is consistent with the joint density of the observation $y$ and $(\x_S, \x'_S)$. We highlight that JBshap resembles on-manifold value function since it respects the data manifold, but can also be considered as off-manifold since it takes an interventional approach similar to off-manifold value functions.

A natural extension of JBshap is to average over all possible baseline values $\x'$, and we obtain RJBshap:

\begin{mydef} \label{def:rJBshap}
(RJBshap): \vspace{-2mm}
\begin{align*} 
    \ \ \ \ \ \ \ \ \  \f^{RJ}_{\x,\p,\prob}(S) = \int_{\x'_{\Bar{S}}} [\p(\x_S; \x'_{\Bar{S}})\prob(\x_S; \x'_{\Bar{S}})\prob_b(\x')] d\x'_{\Bar{S}},
\end{align*}
\end{mydef}
\vspace{-2mm}

where $\prob_b(\cdot)$ is the prior probability for a set of baseline values. We point out that RJBshap still satisfies the axioms (Lin.), (Sym.), (Dum.), (Null), (Eff.), (Rob), with the slight modification of (Sym.), (Null), (Eff.) by replacing the baseline $\x'$ in JBshap by the random baseline $\x'$ with distribution $\prob_b(\x')$

It is easy to observe that JBshap is equal to Bshap when the function $\p(\x) $ is replaced by $\p(\x) \prob(\x)$ since $\f^J_{\x,\p,\prob}(S) = \f^B_{\x,\p \cdot \prob,\prob}(S)$. The clear difference is that JBshap takes account of the data density $\prob(\cdot)$ and thus will not be effected by off-manifold regions $\x$ since $\prob(\x)$ will be small. We can also relate RJBshap and CES, by noting that CES value function can be written as $\f^C_{\x, \p, \prob}(S) = \E_{\x'_{\Bar{S}} \sim \prob(\x'_{\Bar{S}}| \x_S)} [\p(\x_S; \x'_{\Bar{S}})] = \int_{\x'_{\Bar{S}} }\p(\x_S; \x'_{\Bar{S}}) p(\x'_{\Bar{S}}; \x_S) / p(\x_S) d \x'_{\Bar{S}}$. When the baseline follows an uniform solution for all possible values (such that $\prob_b$ is a constant), CES corresponds to RJBshap divided by $p(\x_S)/C_0$, where $C_0$ is the constant value of $\prob_b$. However, we point out that the calculation of $p(\x_S)$ requires marginalizing over $|\Bar{S}|$ variables to obtain $p(\x_S)$, which is computationally difficult. Thus, RJBshap is much more computational tractable compared to CES. By not dividing over $p(\x_S)$, both JBshap and RJBshap are strong T-robust to off-manifold manipulations, in contrast to CES.

\begin{figure*}[h] 
    \vspace{0mm}
        \centering     
    \includegraphics[width=0.99 \linewidth]{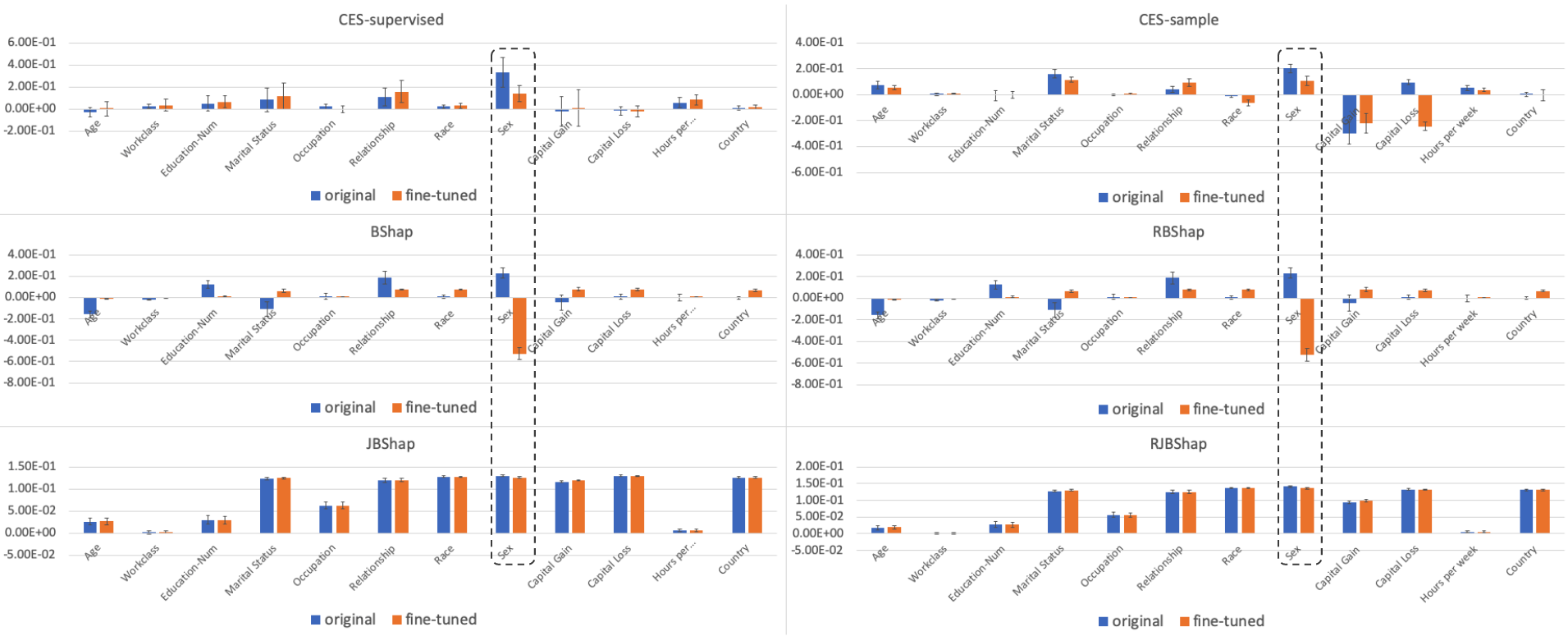}
    \vspace{0mm}
    \caption{Global Shapley values for different value functions on the UCI Adult dataset; with two bars for each feature: on an original model (blue, left) and a fine-tuned model (orange, right). The importance for ``sex'' feature (which is boxed) of RBshap, Bshap, and CES is significantly reduced after the fine-tuning, while the importance for ``sex'' feature of JBshap is almost unchanged.}
    \label{fig:hide_unfair}
    \vspace{-3mm}
\end{figure*}
\vspace{-1mm}
\subsection{Estimating $\prob(\x)$}
\label{sec:px}
\vspace{-1mm}
In our discussion of off and on-manifold value functions, we discussed that the key caveat of on-manifold value function of CES in particular is that it requires expectation (or sampling) with respect to the conditional distributions  $\prob(\x'_{\Bar{S}}| \x_S)$. But if JBshap also requires computing the joint distribution, is it more computational efficient compared to CES at all?

We emphasize that JBshap is much more tractable computationally, since instead of the difficult conditional expectations, we only need to compute the much more amenable \emph{joint density} $\prob(\x'_{\Bar{S}}; \x_S)$. Moreover, we only need estimates of the data density on the space $\x'_{\Bar{S}}; \x_S$ instead of the entire domain $\R^d$. Towards this, we could use noise-contrastive estimation \citep{gutmann2010noise} to estimate the data density on space in the form $\x'_{\Bar{S}}; \x_S$. The idea of noise-contrastive estimation is to estimate the data density of an unknown distribution by comparing it to a self-specified noise distribution. In this case, the event space where we need to query the data density $\prob(\x'_{\Bar{S}}; \x_S)$ is the subspace $[\x'_{\Bar{S}}; \x_S]$ for all possible $\x$ and $S$. We thus specify a self specified noise $Q$ to have an constant probability for every possible $[\x'_{\Bar{S}}; \x_S]$.

To estimate $\prob(\x'_{\Bar{S}}; \x_S)$, we can then train an Out-of-Distribution (OOD) classifier model $\text{OOD}(\x)$ that outputs $1$ when the input $\x \in D$ comes from the true data and outputs $0$ when the input $\x \in Q$ is self-generated noise, and control the input to the training of the OOD classifier to have balanced true and noise data. When the OOD classifier converges, $\text{OOD}(\x) = \prob(\x \in D| \x) = \frac{\prob(\x | \x \in D)}{\prob(\x | \x \in D)+ \prob(\x | \x\in Q)}$. By simple algebra, we can then get $\prob(\x | \x \in D) = \prob(\x | \x \in Q) \times \frac{\text{OOD}(\x)}{1-\text{OOD}(\x)}$.
We emphasize that since we do not require sampling from $\prob(\x'_{\Bar{S}}; \x_S)$, and the subspace of $\x'_{\Bar{S}}; \x_S$ is much more tractable compared to the real space $\R^d$, and the estimation of $\prob(\x'_{\Bar{S}}; \x_S)$ is a much easier problem than estimating $\prob(\x)$ for general $\x$. And estimating even the full joint $\prob(\x)$, is much easier than computing the conditional expectations $\prob(\x'_{\Bar{S}}| \x_S)$ (to provide a sense of the difficulty of the latter: even for simpler probabilistic graphical models, it is always tractable to compute the unnormalized joint density, while intractable in general to compute conditional marginals.) 

While the calculation of JBshap only requires learning $\prob(\x)$ by training an OOD classifier, the calculation of CES either requires learning $\prob(\x)$ and sampling from it for each query (CES-Sample), or retraining a new model that estimates the conditioned value (CES-Supervised). Thus, the computation for calculating JBshap is much more affordable compared to the calculation of CES value functions, as we verify in the experiments.

\vspace{-2mm}

\section{Experiments}
\label{sec:experiment}

\vspace{-2mm}

\begin{figure*}[h!] 
        \centering     
    \includegraphics[width=0.95 \linewidth]{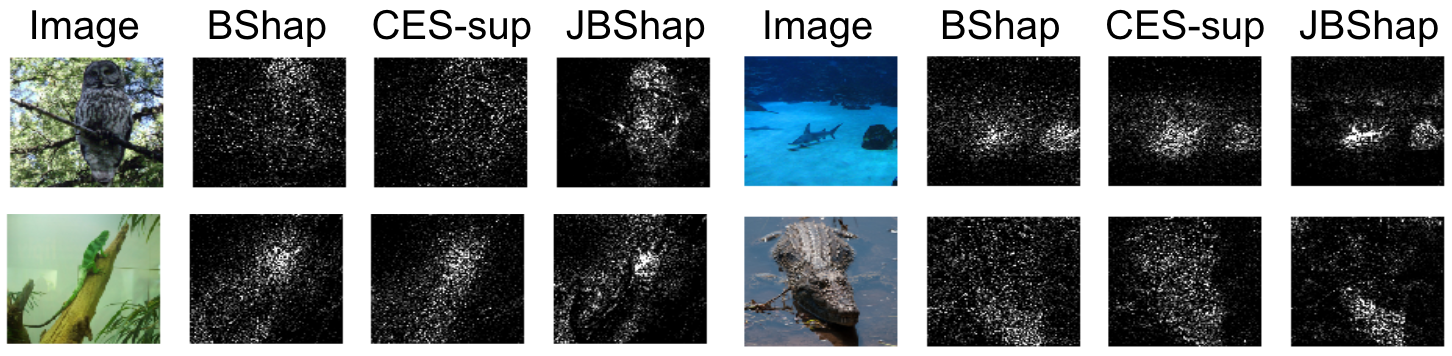}
    \vspace{-5mm}
    \caption{Visualization of Shapley values for JBshap, Bshap, CES-Supervised on Imagenet.}
    \label{fig:imagenet_vis}
    \vspace{-3mm}
\end{figure*}

\subsection{Robustness to off-manifold manipulation.}

Recall that in Sec. \ref{sec:sensitivity}, we have shown that both Bshap and CES are not strong $T-$robust, and thus are prone to manipulations in low-density regions.
To evaluate the robustness to off-manifold manipulations in practice, we perform an empirical study on the UCI Census Income data dataset \citep{Dua:2019} for value functions Bshap, RBshap, JBshap, RJBshap, CES-Supervised, CES-Sample. Given a pretrained biased model, we want to test whether we are able to hide the dependency of a biased feature in the Shapley value by fine-tuning the model only on off-manifold regions.

We trained a four-layer neural network to predict whether an individual’s income exceeds \$50k based on demographic features in the data based on the UCI income Data. To create a biased model, we add $0.1$ times the feature value of ``sex'' to the individual’s income following the setting of \cite{slack2020fooling}. We then train a ``fine-tuned network'' such that the dependency of the feature ``sex'' is set to the negative off the data manifold (while it is positive on the data manifold), while making the same prediction as the original biased model on over $99 \%$ of the testing points. To validate that the bias is not hidden in the $1 \%$ testing point with different predictions, we find that the average difference (L1 Loss) in the prediction probability for data where the fine-tuned model and the original model disagrees (and agrees) is $0.07$ ($0.01$), demonstrating that the model prediction for fine-tuned model and original model are close for all test data points.

For CES-Supervised, we train a ``fine-tuned network'' for the surrogate model $g$, with the constraint that the fine-tuned network behaves similar to the original surrogate model on the on-manifold points and has a similar empirical MSE loss as the original surrogate model (validation MSE loss for the fine-tuned model decreased from $0.0278 \pm 0.001$ to $0.0277  \pm 0.001$,  while training MSE Loss increased from $0.0136 \pm 0.001$ to $0.0139 \pm 0.001$). To determine off-manifold data, we train an OOD detector and use noise-contrastive estimation to obtain $p(\x)$ to calculate JBshap and CES-Sample. For numerical stability, we clip the output the of the OOD classifier between the range $[0.01, 0.99]$. For CES-Sample, we sample 100 points per value function with $\prob(\x)$. We then average the Shapley value based on different value functions on 100 testing points with feature of ``sex'' having value 1 to obtain the global Shapley value. We do 5 separate runs with different data splits (via random seeds), we report the global Shapley value for different value functions in the top figure with error bars representing the standard deviation. We normalize the global Shapley value such that the sum of absolute value of Shapley value for all features sum up to one.

\begin{figure*}[t!] 
    \centering     
    \includegraphics[width=0.7 \linewidth]{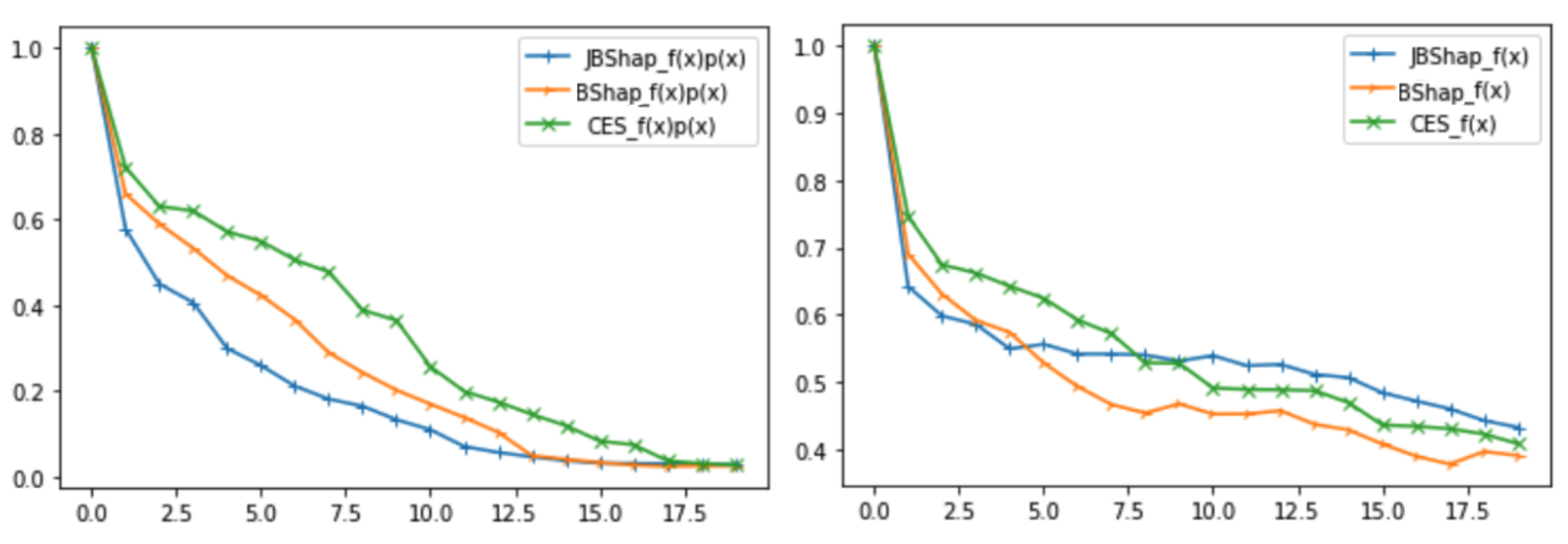}
    \label{fig:delete}
    \vspace{-5mm}
    \caption{Deletion curve for JBShap, BShap, and CES-supervised on Imagenet for joint density (left) and model output (right).}
    \vspace{-5mm}
\end{figure*}

We present our results in  Figure \ref{fig:hide_unfair}; each feature has two bars, each corresponding to the average normalized Shapley values, of the original model (left) and the fine-tuned model (right). We observe that the importance of the feature ``sex'' of the fine-tuned model is strongly reduced for existing value functions Bshap, RBshap and has a negative value for the fine-tuned model. For CES-Sample and CES-Supervised, the respective Shapley value of the feature ``sex'' is not perturbed as severely as Bshap and RBshap, but they are also reduced more than half compared to the original. On the other hand, the Shapley value for JBshap and RJBshap are only slightly reduced for the fine-tuned model. This verifies our theoretical analysis that JBshap is robust to off-manifold manipulations, in contrast to existing value functions. We also note that the Shapley value of JBshap for all features have standard deviation less than $0.01$, demonstrating the robustness of JBshap over different OOD classifiers and data splits.

\vspace{-3mm}

\subsection{Visualization on High Dimensional Data.}
In this section, we visualize the Shapley values for different value functions on image data. We perform experiments on Imagenet subset with the first 50 classes (the total number of classes is $1000$). We first fine-tune a Resnet-18 model that reaches $80 \%$ top-1 accuracy as our base model. We then train a masked surrogate model with the same architecture to calculate CES-Supervised, and we train an OOD classifier to estimate $\prob(\x)$ by noise contrastive estimation to calculate JBshap. To calculate the Shapley value, we use the permutation sampling approach introduced in \citet{castro2009polynomial} with $10$ permutations per image. We compare the Shapley values from different value functions: Bshap, JBshap, CES-Supervised. Note that we omit other variants of value function since they require more than one function pass during the calculation of the value function, which will make the computation infeasible on high-dimension. The training of the surrogate model takes around $2976$ seconds, while the training of the OOD classifier only take around $74$ seconds. The cost of the training of the surrogate model will become even more costly if the model is trained on the full set of imagenet, not just a subset of imagenet. Thus, the additional training time for JBshap is much lower than the additional training time for CES-Supervised. The Shapley value computation for each image takes around $1800$ seconds, which is equal for Bshap, JBshap, and CES-supervised. All computations are done on a single 1080-Ti GPU. The visualization results are shown in Figure~\ref{fig:imagenet_vis}, and we can observe that while the Shapley value for different value functions all focus on the object in the image, the Shapley value for JBshap concentrates on a smaller part of the image. For instance, in the bottom right image, Bshap and CES-supervised focuses on the whole alligator, while JBshap focuses more specifically on the head on the alligator. One possible reason is that Bshap and CES-supervised captures pixels that contributes to the model prediction, while JBshap focuses on pixels that contributes to both the data density of the image and the model prediction.

\paragraph{Quantitative evaluation on ImageNet}

We perform an evaluation for the Shapley values based on different variants of value function, where we choose the well-known deletion evaluation criteria for our results in ImageNet over the average of 50 data points, where we remove the top pixels by Shapley value and evaluate how the function decreases (lower is better)~\citep{samek2016evaluating, arras2019evaluating} in Fig.~\ref{fig:delete}, where we show the resulting joint density $f(x)p(x)$ in the left and model value $f(x)$ in the right after removing top pixels, where the x-axis is the percent of pixels. The area under curve (AUC) (where smaller is better) of $f(x)p(x)$ for JBShap, Bshap, CES is $0.207,0.27, 0.348$ respectively, and the AUC for $f(x)$ for JBShap, Bshap, CES is $0.549,0.504,0.556$ respectively. Thus, we hypothesize that JBShap highlights features that affect both $f(x)$ and $p(x)$, and Bshap highlights pixels that only affect $f(x)$. We also report the rank correlation version of sensitivity-n~\cite{ancona2017unified}, which measures the correlation between the function drop and sum of Shapley values for random sets of pixels, where the random sets of pixels is set to $1-20$ percent of pixels, and the average sensitivity-n rank correlation for JBShap, Bshap, CES, are $0.182 \pm 0.007, 0.132 \pm 0.004, 0.119 \pm 0.01$ respectively.

\vspace{-3mm}
\section{Conclusion}
\vspace{0mm}
In this work, we discussed the issues of interventional and conditional value functions, including different implementation variants of conditional value functions. To address these issues, we propose JBshap, which uniquely satisfies a set of axioms considering both the model and data manifold, while being robust to manipulation and computationally efficient.

\paragraph{Ackowledgement}
We acknowledge the support of ONR via N000141812861, and NSF via IIS-1909816.

\newpage

\bibliographystyle{unsrtnat} 
\bibliography{interpret_ref.bib}

\clearpage
\appendix

\section{Issues for CES Versions}

\subsection{Sparsity Issue For CES-Empirical and CES-Supervised}\label{sec:ces_emp_issues}

We discuss an issue for CES-Empirical and CES-Supervised in this section with a concrete example. 

\begin{myex} \label{ex:ces_2}
Consider a model where $\p(\x) = 10 \x_1 - 10 \x_2,$ where there are three candidate features $\x_1, \x_2, \x_3$. The data distribution is generated from a multivariate Gaussian $\mathcal{N}(0,I)$. However, we do not know the true distribution of the data, and we can only model the distribution by a mixture Gaussian model where each data point is the center of a Gaussian ball. Suppose we are to explain the outcome of a data point $(3,-3,10)$. Here, $\x_3$ is clearly an outlier value, which is way larger than other data points. Therefore, the value function of $S = \x_3$ for CES is equal to $\f_{\x, \p, \prob}(\{\x_3\}) = \E_{\x'_{\Bar{S}} \sim p(\x'_{\Bar{S}}| \x_3 = 10)} [\p(\x_3 = 10; \x'_{\Bar{S}})]$ will be very close to 60. On the other hand, the value of feature 1 is not an outlier, the value function of $S = \x_1$ for CES will be close to $\f_{\x, \p, \prob}(\{\x_1\}) = \E_{\x'_{\Bar{S}} \sim p(\x'_{\Bar{S}}| \x_1 = 3)} [\p(\x_1 = 3; \x'_{\Bar{S}})] = 30$. It is easy to see that for this data point, Shapley value for CES will assign a higher importance value to feature 3 than feature 1 due to sparsity of outlier values, even if feature 3 is not related to the model at all.
\end{myex}
\vspace{-2mm}

\begin{figure*}[t]
\centering
{
\includegraphics[width=0.8\textwidth]{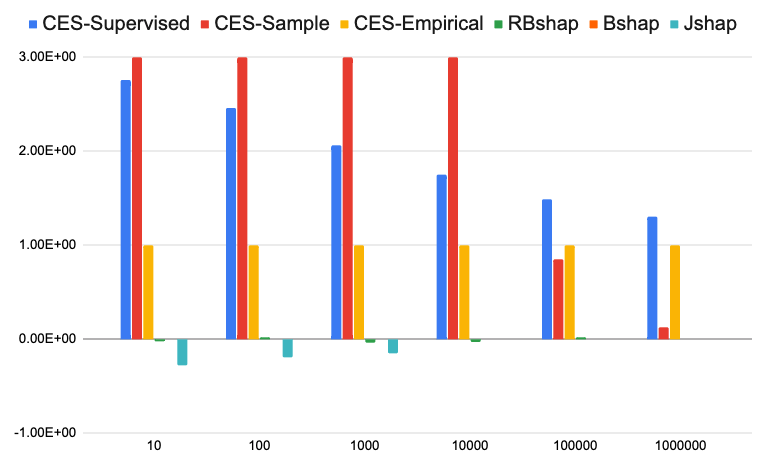}}%
\caption{Relative importance ratio of feature 3 with respect to the average importance of feature 1 and feature 2, the ideal value should be $\leq 0$. We show the relative important to be 3 if the actual value is larger than 3 for better visualization.}
\label{fig:outlier_all}
\end{figure*}

Since the true probability distribution is not known in this case, we use the smoothed empirical distribution $P^{SE}(\x) = \sum_{i=1}^m \frac{1}{m} \mathcal{N}(\x_i,I) (\x)$ to estimate the distribution for JBshap and CES-Sample. We consider CES-Emp, CES-Sample, CES-supervised, Bshap, RBshap, JBshap in this example. For Bshap and JBshap, we let the baseline be the average value of the dataset. For CES-supervised, we learn a KNN regressor with a Gaussian kernel as the predictor. We report the relative importance of feature 3 compared to the average importance of feature 1 and feature 2 for each Shapley value in Figure \ref{fig:outlier_all}, where the ideal value should be $\leq 0$. The X-axis is the number of the data in the dataset, which effects CES-Sample, JBshap by the smoothed empirical distribution, effects CES-Empirical by the empirical distribution, and effects CES-Supervised by the training of the supervised value function.

We observe that CES-Empirical gives equal contribution to feature 3 compared to feature 1 and feature 2 due to the sparsity issue (see Remark 3.2 in \citet{sundararajan2019many}). For both CES-Sample and CES-Supervised, the relative importance of feature 3 is still higher than 1 with around 1000 data points. We further observe that the relative importance of feature 3 of CES-Sample decreases to near $0$ when more than 10000 data points are in the dataset. However, even with 100,000 data points in the dataset, the relative importance of feature 3 is still larger than 1 for CES-supervised (but converging to 1). This is not surprising since in Proposition \ref{thm:ces-sup} we showed that CES-Supervised converges to CES-Empirical, which suffers from the sparsity issue. While CES-Sample may need more data points to obtain the correct Shapley value, CES-supervised do not converge to the correct value even with $100,0000$ data points due to the sparsity issue.

On the other hand, JBshap, Bshap, RBshap gives a low importance value for feature 3 even with only 10 data points. Thus, CES-Sample should be preferred over CES-Supervised and CES-Empirical when the data is continuous and the sparsity issue may become a problem.

\subsection{Issue for CES-Masked-VAE} \label{sec:cesvae_issue}
\begin{figure}[h!]
\centering
{
\includegraphics[width=0.5\textwidth]{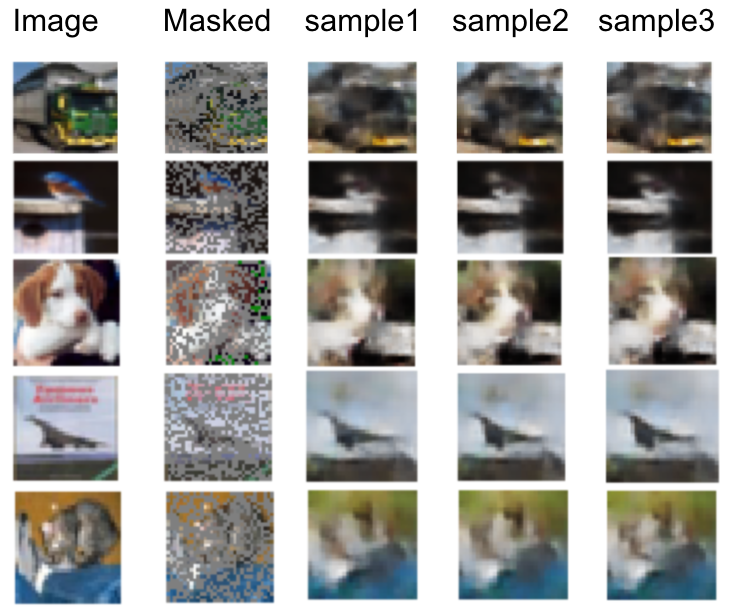}}%
\caption{Sampling result from masked VAE.}
\label{fig:cifarvae}
\end{figure} 

In Sec. \ref{sec:cessample}, we mentioned that one version of CES sample is proposed by \cite{frye2020shapley} to used a masked variational autoencoder, which we call CES-Masked-VAE. We briefly explain the proposed method and a major flaw of the method.

\paragraph{CES-Masked-VAE} 

Suppose we have a density model that computes  $\prob(\x|\x_S)$ for all possible subsets $S \subseteq [d]$;     \citet{frye2020shapley} suggest using masked VAEs to do so. One can then sample $\x'_{\Bar{S}}$ from $\prob(\x'_{\Bar{S}}| \x_S)$, and calculate $\E_{\x'_{\Bar{S}} \sim \prob(\x'_{\Bar{S}}| \x_S)} [\p(\x_S; \x'_{\Bar{S}})]$ via Monte-Carlo sampling. We use masked VAEs as a density model in our experiments, following the suggestion of \citet{frye2020shapley}, and hence term this approach CES-Masked-VAE.

\paragraph{Unfaithfulness of Masked CES-Masked-VAE to coalition constraints.}

The CES-Masked-VAE approach in \citet{frye2020shapley} trains a masked encoder $r_\Psi(z|\x_S)$ that aims to agree with the full encoder $q_\phi(z|\x)$ by minimizing the KL-divergence between them. However, one issue is that this does not guarantee that the masked encoder $r_\Psi(z|\x_S)$ will have support in latent spaces corresponding to $\x'$ where $\x'_S = \x_S$. Thus, sampling from $\prob(\x'|\x_S) = \int dz r_\Psi(z|\x_S) \prob_\theta(\x'|z)$ may obtain $\x'$ which does not satisfy $\x'_S = \x_S$, which is unfaithful to the conditioned value. To verify this, we trained a CES-Masked-VAE on Cifar-10 with methods specified in \citet{frye2020shapley}, and sample $\x'$ from the Masked VAE estimate of $\prob(\x'|\x_S)$. We show some visualization results in Fig. \ref{fig:cifarvae}. We can observe that each image, the sampled images disagree with the masked image on parts that are not masked. For instance, the color of the truck head in image 1, the color of the bird in image 2, the right ear of the dog in image 3, the caption in image 4, and the color of the background in image 5. This verifies that the masked VAE approach does not generate samples from true $\prob(\x'|\x_S)$, as the samples $\x'_S$ should be equal to the original image $\x$ on pixels that are not masked.

\section{More Visualization Results }
\begin{figure*}[t!]
\centering
{
\includegraphics[width=0.98\textwidth]{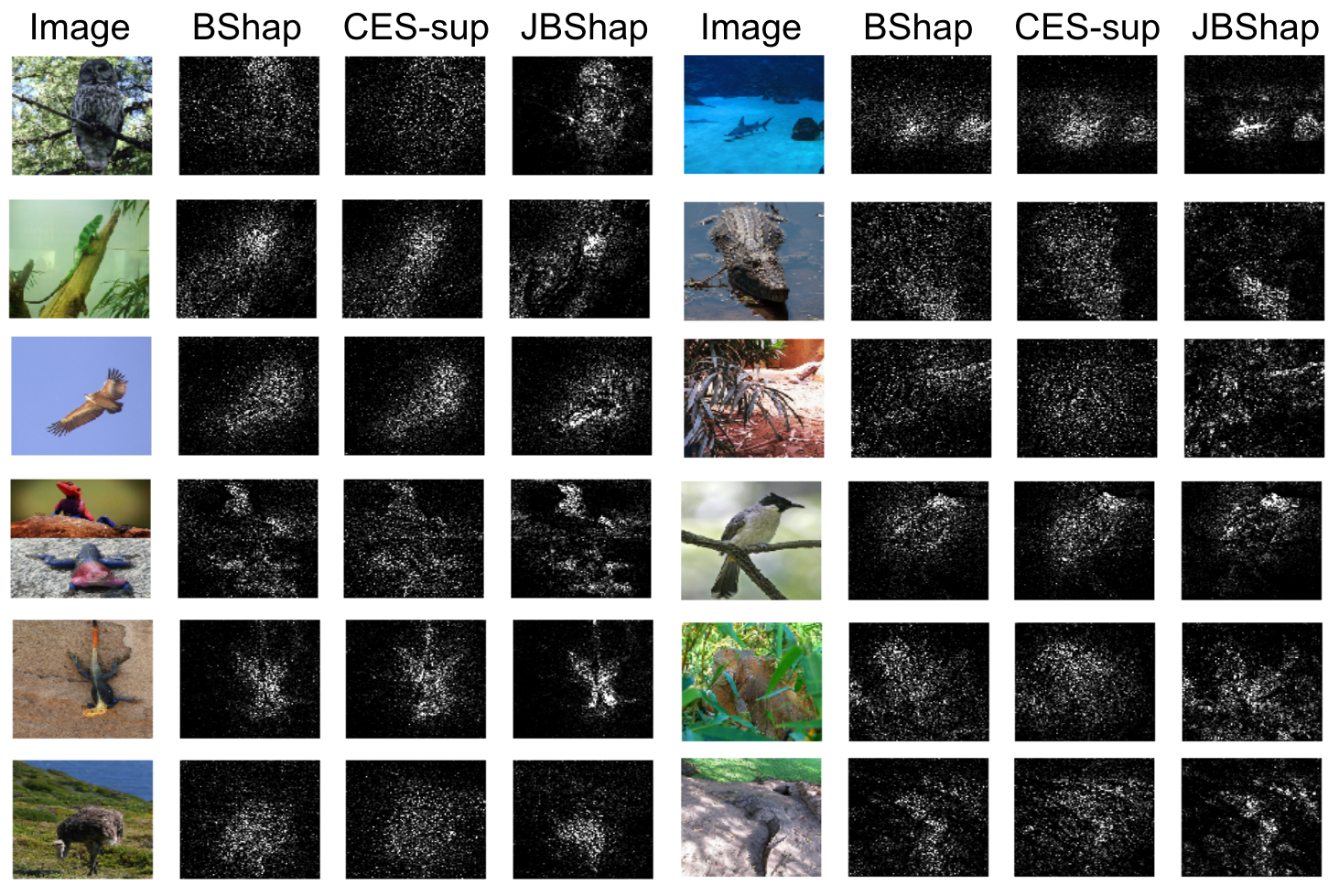}}%
\caption{More Visualization result on Imagenet.}
\label{fig:more_vis}
\end{figure*}

In this section, we include more visualization results for Shapley values of Imagenet images. We show the visualization results for Bshap, JBshap, CES-Masked-VAE, CES, and RJBshap with different testing images in Fig. \ref{fig:more_vis}.

\section{Translative Relation For Shapley value Axioms} \label{sec:translative}

Recall the set of axioms we introduced in Sec. \ref{sec:new_axiom}, we rename their abbreviations here for simpler presentations.

Lin. in main text is renamed as (L-DS), Sym. in main text is renamed as (S-DS), Dum. in main text is renamed as (D-DS), Null in main text is renamed as (N-DS), Eff. in main text is renamed as (E-DS), Set. in main text is renamed as (Set-DS), and the renaming makes the axioms easier to distinguish to other versions of the respective axioms.

\begin{enumerate}
	\item {\bf Linearity (over functions and distributions) (L-DS)}: For any functions $\p, \p_1,\p_2, \prob, \prob_1, \prob_2$, $\alpha_1 \f_{\x,\p_1,\prob}(S) + \alpha_2 \f_{\x,\p_2,\prob}(S) = \f_{\x,\alpha_1 \p_1+\alpha_2 \p_2, \prob}(S)$, for $\alpha_1,\alpha_2 \in \mathbb{R}$ and $\alpha_1 \f_{\x,\p,\alpha_1\prob_1}(S) + \alpha_2 \f_{\x,\p,\prob_1}(S) = \f_{\x,\p, \alpha_1\prob_1 + \alpha_2\prob_2}(S)$ for $\alpha_1, \alpha_2 \ge 0; \alpha_1 + \alpha_2 = 1$.
	\item {\bf Symmetry (over functions and distributions) (S-DS)}: for all $i,j \not \subseteq S$,
$\p$, $\prob$, $\x$, $\x'$ all being symmetric in the $i$-th and $j$-th dimensions implies  $\f_{\x,\p,\prob}(i\cup S) = \f_{\x,\p,\prob}(j\cup S)$.
	\item {\bf Dummy Player (over functions and distributions) (D-DS)}: $\p, \prob$ being invariant in the $i$-th dimension implies $\f_{\x,\p,\prob}(S) = \f_{\x,\p,\prob}(i\cup S)$.
	
	\item {\bf Null Player (over functions and distributions) (N-DS)}:  if $\p', \prob'$ satisfies $ \p(\x') = \p'(\x'), \prob(\x') = \prob'(\x') $, then $\f_{\p,\prob,\x}(\text{\O}) = \f_{\p',\prob',\x}(\text{\O})$.
	\item {\bf Efficiency (over functions and distributions) (A-DS)}: $ \f_{\p,\prob,\x}[d] - \f_{\p,\prob,\x}[\text{\O}] = \p(x) \prob(\x) - \p(x') \prob(\x')$.
	\item {\bf Set Relevance (over functions and distributions) (Set-DS)}: If $ \p_1(\x_S, \bar{x}_{\bar{S}})  = \p_2(\x_S, \bar{x}_{\bar{S}})$ and $ \prob_1(\x_S, \bar{x}_{\bar{S}})  = \prob_2(\x_S, \bar{x}_{\bar{S}})$ for any $\bar{x}$, $\f_{\p_1,\prob_1,\x}(S) = \f_{\p_2,\prob_2,\x}(S)$.
	
	\item {\bf Strong-T-Robustness (over functions and distributions) (Rob-DS)}: For any functions $\p_1,\p_2, \prob$, if $\max_\x |\p_1(\x)-\p_2(\x)| \prob(\x)  \leq \epsilon$, then $|\f_{\x,\p_1,\prob}(S) - \f_{\x,\p_2,\prob}(S)| \leq T \epsilon$.
    \vspace{-1.5mm}
\end{enumerate}

We then introduce the following definitions for a set of axioms for a value function. Define the set $[d]:=\{1,2,\ldots,d\}.$ We first start from axioms with only a single input function $f$:
\vspace{0mm}
\begin{enumerate}
	\item {\bf Linearity (over functions) (L-IS)}: $\f_{\x,\p_1,\prob}(S) + \f_{\x,\p_2,\prob}(S) = \f_{\x,\p_1+\p_2, \prob}(S)$.
	\item {\bf Symmetry (over functions) (S-IS)}: for all $i,j \not \subseteq S$,
$\p$, $\x$, all being symmetric in the $i$-th and $j$-th dimensions implies $\f_{\x,\p,\prob}(i\cup S) = \f_{\x,\p,\prob}(j\cup S)$. 
	\item {\bf Dummy Player (over functions) (D-IS)}: $\p$ being invariant in the $i$-th dimension implies $\f_{\x,\p,\prob}(S) = \f_{\x,\p,\prob}(i\cup S)$.
    \item {\bf Null Player (over functions) (N-IS)}:  if $ \p(\x) = \p'(\x)$, then $\f_{\p,\prob,\x}(\text{\O}) = \f_{\p',\prob,\x}(\text{\O})$.
	\item {\bf Efficiency (over functions) (E-IS)}: $ \f_{\p,\prob,\x}[d] - \f_{\p,\prob,\x}[\text{\O}] = \p(x) - \p(x')$.
	\item {\bf Set Relevance (over functions) (Set-IS)}: If $ \p_1(\x_S, \bar{x}_{\bar{S}})  = \p_2(\x_S, \bar{x}_{\bar{S}})$ for any $\bar{x}$, $\f_{\p_1,\prob,\x}(S) = \f_{\p_2,\prob,\x}(S)$.
	\vspace{-2mm}
\end{enumerate}

The difference is that these set of axioms disregard the data distribution $\prob$.

We then define the following axioms between the {\bf S}et function $v$ and {\bf E}xplanation $\phi$ as Axioms-SE. This is a classic set of axioms in game theory and has been widely discussed. See, e.g., the work of \citet{sundararajan2019many, shapley_1988, lundberg2017unified}.

\begin{enumerate}
	\item {\bf Linearity from set to explanation (L-SE)}: $\ex(\f_1) + \ex(\f_2) = \ex(\f_1+\f_2)$ for any two value functions $\f_1$ and $\f_2$.
	\item {\bf Symmetry from set to explanation (S-SE)}: $\f(S \! \cup  i) = \f(S \! \cup  j) \!$ for any $S \subseteq [d] \backslash \{i,j\}$, $\ex_{i}(\f) = \ex_{j}(\f)$.
	\item {\bf Dummy Player from set to explanation (D-SE)}: $\f(S \cup i) = \f(S)$ for any $ S \subseteq [d] \backslash \{i\}$ implies $\ex_{i}(\f) = 0$.
	\item {\bf Efficiency from set to explanation (E-SE)}: $\sum_{i} \ex_i(\f) = \f([d]) - \f(\text{\O})$.
\end{enumerate}

The Shapley value is well-known to be the unique function that satisfies the above four axioms. However, this set of axioms assumes a set function. On the other hand, the machine learning model takes a real-valued input instead of a set.

\begin{figure}
    \centering
    \includegraphics[width = 0.9\linewidth]{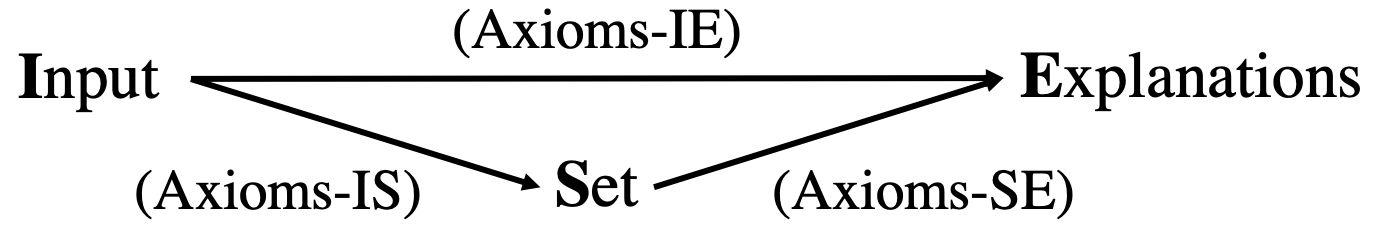}
    \caption{An illustration of the relation between the three sets of axioms. Axioms-IS and Axioms-SE satisfy a transfer property: broadly speaking, a pair $v, \phi$ satisfies Axioms-IE if $v$ satisfies Axioms-IS and $\phi$ satisfies Axioms-SE.}
    \label{fig:axiom-relation}
    \vspace{-6mm}
\end{figure}

Thus, we propose the following set of axioms between the {\bf I}nput functions ($f$ and $p$) and {\bf E}xplanation $\phi$, which we term Axioms-IE. Axioms-IE can be viewed as a parallel to the axioms proposed by \citet{sundararajan2019many}; the main difference is that our axioms take into account the in-data classifier $p$. 
\vspace{-2mm}
\begin{enumerate}
	\item {\bf Linearity from double input to explanation (L-DE)}: $\ex_{\p_1,\prob}(\x) + \ex_{\p_2,\prob}(\x) = \ex_{\p_1+\p_2,\prob}(\x)$ and $\ex_{\p,\prob_1}(\x) + \ex_{\p,\prob_2}(\x) = \ex_{\p,\prob_1 + \prob_2}(\x)$, for any functions $\p, \p_1,\p_2, \prob, \prob_1, \prob_2$.
	\vspace{-1mm}
	\item {\bf Symmetry from  double input to explanation (S-DE)}: if $\p$, $\prob$, $\x', \x$ are symmetric in the $i$-th and $j$-th dimensions, then $\ex_{\p,\prob,i}(\x) = \ex_{\p,\prob,j}(\x)$.
	\vspace{-1mm}
	\item {\bf Dummy Player from double input to explanation (D-DE)}: both $\p, \prob$ being invariant in the $i$-th dimension implies $\ex_{\p,\prob,i}(\x) =0$. In other words, if feature $i$ is not used in the model, then 
	\vspace{-1mm}
	\item {\bf Efficiency from double input to explanation (E-DE)}:  $\sum_i \ex_{\p,\prob,i}(\x) =  \p(x) \prob(\x) - \p(x') \prob(\x')$.
\end{enumerate}
\vspace{-2mm}

We remark that when ignoring $p$, this recovers a version of Axioms-IE proposed by \citet{sundararajan2019many}: when ignoring $p$, it should be quite evident that value functions that depend on $p$ (such as CES and JBshap) would easily fail to satisfy linearity and dummy defined by \cite{sundararajan2019many}. However, we argue that (CD-DE) is the form of dummy player axiom satisfied by CES and (D-DE) is the form of dummy player axiom satisfied by JBshap. This suggests that previous discussions on this topic may need restatements based on our proposed Axioms-IE, which takes into account the dependencies of $p$.

The intricate relation between Axioms-IS, Axioms-SE and Axioms-IE is captured in the following theorem.
\begin{theorem} 
If set function $ \f_{\x,\p,\prob}(\cdot)$ satisfies Axioms-IS, and explanation $\phi_{\x,\p,\prob}(\cdot)$ satisfies Axioms-SE, then $(\f_{\x,\p,\prob}(\cdot), \ex(\f_{\x,\p,\prob}(\cdot)) )$ satisfies Axioms-IE. Similarly, if set function $ \f_{\x,\p,\prob}(\cdot)$ satisfies (L-DS), (S-DS), (D-DS), (M-DS), (N-DS), and (E-DS), and explanation $\phi_{\x,\p,\prob}(\cdot)$ is the Shapley value, then $(\f_{\x,\p,\prob}(\cdot), \ex(\f_{\x,\p,\prob}(\cdot)) )$ satisfies (L-DE), (S-DE), (D-DE), (E-DE).
\label{thm:thm_transfer_1}
\vspace{0mm}
\end{theorem}

Here, theorem \label{thm:thm_transfer} states the translative relationship between set of axioms from input to set, set to explanations, and input to explanations, as shown in Fig. \ref{fig:axiom-relation}.

This further motivates the design of axioms shown in Sec.\ref{sec:new_axiom}.

We remark that these sets of axioms can be useful to explanations not limited to the Shapley value. For instance, the Banzhaf value (SE version) shares a set of axioms with the Shapley value (SE version), together with our proposed (DS version of axioms), we may get resulting (DE) axioms for the Banzhaf value also. However, since this paper is focused to the Shapley value as an explanation, we do not further discuss explanations other than the Shapley value. 

\section{Experiment Details}

\subsection{Hiding Unfairness with Noise contrastive estimation}

For the hiding unfairness experiment, we use the noise contrastive estimation to estimate the data density $\prob$. We build the OOD classifier, we let the noise be $\x_S \sim \prob(\x|\x \in Q)$ where $\x \sim \prob^E(\x)$, $S \sim \text{uniform over all set}$, where the noise is masked image which is equally likely for any available masking. Therefore, for any two $\x_S$ in the noise distribution, they will have exactly the same pdf. We then train a OOD classifier on equal number of true data $D$ and noise $Q$, using the binary cross entropy loss. The resulting $\text{OOD}(\x)$ will be equal to $\prob(\x \in D|\x) = \frac{\prob(\x | \x \in D)}{\prob(\x | \x \in D)+ \prob(\x | \x\in Q)}$ if the OOD classifier is trained on balanced noise and true data. By some calculation we get $\prob(\x | \x \in D) = \prob(\x | \x \in Q) \times \frac{\text{OOD}(\x)}{1-\text{OOD}(\x)}$. Since $\prob(\x | \x \in Q)$ is equal for any possible point in $Q$, we may treat it as a constant, and thus $\prob(\x | \x \in D) \propto \frac{\text{OOD}(\x)}{1-\text{OOD}(\x)}$.

By Thm.2 in \citet{gutmann2010noise}, since $\prob(\x | \x \in Q)$ is non-zero whenever $\prob(\x | \x \in D)$ is non-zero, and if additionally $\int g(\x) g(\x) P(\x) \prob_d(\x) d\x$ has full rank, where $P(\x) = 1- OOD(\x)$ and $g(\x) = \nabla_\theta \prob(\x| \x \in D, \theta)$, then $\prob(\x|\x \in D) \rightarrow \prob(\x)$, such that our probability estimate $\prob(\x|\x \in D)$ converges to the true density $\prob(\x)$ in probability. Note that we only need a scaled version of $\prob(\x)$, so we also scale $\prob(\x)$ to $[0,1]$ for simplicity.

We normalize all features to have 0 mean an unit variance, and we thus choose baseline value to be 0 for Bshap, JBshap, RBshap, RJBshap. We use two layer NNs for the original model and the fine-tuned model $f'$, and use a three layer NN to model OOD($\x$). Recall that we have added 0.1 times the feature value of ``sex'' to the income value to make the model biased. To create a fine-tuned model such that the Shapley value is negative, we add an additional loss to enforce $f'$ (the fine-tuned model) is negatively correlated to the feature value of ``sex'' on off-manifold regions. To train the fine-tuned model, we have two losses, the first is to enforce $f'$ is close to the original model $f$ on the data manifold (points in the original dataset). The second is to enforce the prediction of $f'$ to be negatively correlated to the value of ``sex'' feature only on regions with the scaled $\prob(\x)$ is smaller than $0.1$. We use squared error for both losses and train the model with a weight $1:100$, while the fine-tuned model still predicts the same result with the original model on over $98$ percent of the data.

To train the fine-tuned surrogate model for CES-Supervised, we also have two losses. the first is to enforce $g'$ (the fune-tuned surrogate model) is close to the original surrogate model $g$ on the data manifold (points in the original dataset). The second is to enforce the prediction of $f'$ to be negatively correlated to the value of ``sex'' feature only on regions with the scaled $\prob(\x)$ is smaller than $0.1$. We use squared error for both losses and train the model with a weight $1:1$, while the fine-tuned model still predicts the same result with the original model on over $99$ percent of the data.

We argue why the surrogate model learned for CES-Supervised can be fine-tuned in this setting. Note that CES-supervised learns a surrogate model based on the MSE loss. However, during the training of the surrogate model, some other losses (such as data augmentation and weight regularization) would have to be added to guarantee convergences on more complicated dataset such as imagenet. Therefore, it would be difficult if not impossible to know whether the learned surrogate model is fine-tuned or not, and the only indicator is the MSE loss. As we show in the main text, the fine-tuned surrogate model has MSE loss very close to the original surrogate model (even lower), and thus it would be difficult to differentiate whether the surrogate model is fine-tuned or not. Our setting assumes that training of the model is a process that the user cannot completely control, and since CES-supervised requires a full retraining of the surrogate model, it is still prone to perturbations.

Since there are only $12$ features, we calculate the exact Shapley value by the formula regression formula in \citet{ancona2017unified}. For the randomized version RBshap, RJBshap, CES-Sample, we sample $10$ points per value function. To report the global Shapley value, we sum the Shapley value over all first $100$ data points whose sex feature is originally ``0'' (before normalization).

\subsection{Visualizing on Imagenet}

We choose the first 50 class in imagenet to reduce the number of training examples (and thus the training cost is reduced) so that CES-supervised will be possible to compute.

For visualization, we mask out the bottom 30 percent of the pixels with the least L1-norm of Shapley value, and scale Shapley value for all pixels to $[0,1]$. We use the softmax output of the target class of the CNN model as the function $\p$. For RJBshap, we let the baseline distribution to be a constant vector with all elements in the vector to range uniformly between $[0,1]$.

We now introduce the sampling method for calculating JBShap on high-dimensional image data. We observe that due to the multiplication of data density $\prob(\cdot)$, the value function of $\f_{\x, \p, \prob}(S)$ is very small when |S| is much smaller than the total number of features. The sampling track proposed by \citet{castro2009polynomial} calculates the Shapley value for feature i as $\E_{S}[\f(S \cup i) - \f(S)],$ where we first sample a permutation of all features, and $S$ is the set of all features that occurred before $i$ in the permutation (if $i$ is sampled first $S = \emptyset$). We slightly modify the sampling by setting $\f(S) = 0$ whenever $|S|< 0.8\times d$, where $d$ is the total features of the model. Thus, whenever the feature $i$ occurs in the first $80 \%$ of the features in the sampled permutation, we use the value $0$ instead of calculating $\f(S \cup i) - \f(S)$. This sampling trick is only applicable to JBShap since $\prob(x)$ is often small when many pixels are set to the baseline value.

\subsection{CES-Masked-VAE}
For the training of the masked VAE and VAE model, we used 4 convolutional layers with 256 filters followed by a dense layer each for the encoder, masked encoder, and the decoder. The latent dimension of $z$ is $256$, and the VAE modeled is trained $500$ epochs and $\beta$ is set to 1, following the structure of \citet{frye2020shapley}. For the reported difference $\E_{\x' \sim \prob(\x'|\x_S)}[\frac{1}{|S|}\|\x_S - \x'_S\|_1]$ and $\E_{\x' \sim \prob(\x'|\x_S)}[\frac{1}{|\bar{S}|}\|\x_{\bar{S}} - \x'_{\bar{S}}\|_1]$, we average over 10 runs, where each runs contain $10000$ testing images each with one sampled mask $S$, and for each $\x_S$ we samples $\x' \sim \prob(\x'|\x_S)$ 20 times, and report the average of $[\frac{1}{|S|}\|\x_S - \x'_S\|_1]$ and $[\frac{1}{|\bar{S}|}\|\x_{\bar{S}} - \x'_{\bar{S}}\|_1]$. 

\subsection{CES-supervised}
We trained the CES-supervised by modifying the training code of regular Imagenet by adding a noise layer in the data generation process to mask the image randomly, and change the training target to the output of a pre-trained model on the unmasked image, and adopt the MSE loss. We follow the normal training of the imagenet with a resnet-18 models, and we trained with 20 epochs with batch size 256, a single GPU, and learning rate $0.01$. 

To evaluate the True MSE loss and the empirical MSE loss, we randomly sample 100 $\x_S$. For each $\x_S$, we calculate $[(\p(\x) -\f^{CS}_{\x_S, \p, \prob})^2]$ for the empirical MSE loss for $\f^{CS}_{\x_S, \p, \prob}$, and $\E_{\x \sim \prob(\x|\x_S)}[(\p(\x) -\f^{CS}_{\x_S, \p, \prob})^2]$ for the true MSE loss for $\f^{CS}_{\x_S, \p, \prob}$, where sampling from $\x \sim \prob(\x|\x_S)$ is done by importance sampling with 100 samples.

\section{Proof} \label{sec:proof}
\subsection{Proof of Proposition \ref{thm:ces-sup}}
\begin{proof}
We consider $\f^{CS}_{\x_S, \p, \prob^E}$ with a fix $\x_S, \p$, and plug in $\prob^E(\x) = \frac{1}{m} \mathbbm{1}[\x \text{ in dataset}]$. Eq. \eqref{eq:CES_supervised} can be rewritten as 
\begin{align}\label{eq:ces_supervised_1}
    \E_{S\sim \text{shapley}}[\sum_{i=1}^m (\p(\x^i) -\f^{CS}_{\x_S, \p, \prob})^2].
\end{align}

Since we consider  $\f^{CS}_{\x_S, \p, \prob^E}$ with $\x_S, \p$ fixed, we can take the first order condition on $\f^{CS}_{\x_S, \p, \prob^E}$ with a fix $\x_S, \p$ in Eq. \eqref{eq:ces_supervised_1} since it is convex, we then obtain the optimal $\f^{CS*}_{\x_S, \p, \prob}$ should satisfy:

\begin{align}\label{eq:ces_supervised_2}
    -\prob_{\text{shapley}}(S)[\sum_{i=1}^m 2\mathbbm{1}[\x^i_S = \x_S](\p(\x_i) -\f^{CS*}_{\x_S, \p, \prob})] = 0.
\end{align}

Thus, we obtain 
\begin{align}\label{eq:ces_supervised_3}
    \f^{CS*}_{\x_S, \p, \prob} = \frac{\sum_{i=1}^m \mathbbm{1}[\x_S^i=\x_S] \p(\x^i)}{\sum_{i=1}^m \mathbbm{1}[\x_S^i=\x_S]} = f^{CE}_{\x, \p, \prob^E}(S).
\end{align}
\end{proof}

\subsection{Proof of Proposition  \ref{prop:notstrong}}
\begin{proof}
Assume that Bshap and CES are strong $T\text{-robust}$ with some $T = T_1 >0$. Consider a model $\p_1(\x) = \x_1,$ with two candidate features $\x_1, \x_2$. We assume the features follow standard Gaussian $\mathcal{N}(0,I)$ distributions. Suppose we are to explain a data point $\x = (0,T_0)$.  We choose $T_0$ such that $T_0 > \text{pdf}^{-1}(\frac{1}{T1})$, and thus $\prob(\x = (0,T_0)) < \prob(\x_2 = T_0) <\frac{1}{T1}$. 

It is clear that $\f^{C}_{\x, \p_1, \prob}(\{\x_2\}) = \f^{B}_{\x, \p_1, \prob}(\{\x_2\}) = 0$ since the model $\p_1$ is unrelated to $\x_2$. However, if we perturb the model by setting $\p_2(\x) = \x_1 + C \mathbbm{1}[\x_2 = T_0]$, $\f^{C}_{\x, \p_2, \prob}(\{\x_2\}) = \f^{B}_{\x, \p_2, \prob}(\{\x_2\}) = C$. $\max_\x |\p_1(\x)-\p_2(\x)| \prob(\x) < \frac{C}{T_1}$, but $|\f^B_{\x,\p_1,\prob}(S) - \f^B_{\x,\p_2,\prob}(S)| = C$ and $|\f^C_{\x,\p_1,\prob}(S) - \f^C_{\x,\p_2,\prob}(S)| = C$. Thus, Bshap and CES are not T-robust for any $T_1>0$.

\end{proof}

\subsection{Proof of Theorem. \ref{thm:JBshap-uniqueness}}
In this section, we provide a proof of Theorem \ref{thm:JBshap-uniqueness}. We first start with the first part: 
If $\f_{\x,\p,\prob}(S)$ satisfies (Lin.), (Sym.), (Dum.), (Null), (Eff.), (Set.), (Rob.) axioms implies $\f_{\x,\p,\prob}(S) =  \p(\x_S;\x'_{\Bar{S}})\prob(\x_S;\x'_{\Bar{S}})$.

We begin with the following proposition, which characterizes the dependency of $v_{\x,f,p}(S)$ on $f, p, \x_S, S$.
\begin{myprops} \label{prop:step2}
If (D-DS), (N-DS), (SET-DS) are all satisfied for input function to set function, then we have $\f_{\x,\p,\prob}(S)$ only depends on $\p(\x_S;\x'_{\Bar{S}}),\prob(\x_S;\x'_{\Bar{S}})$. That is, if for some $\p_1, \p_2, \prob_1, \prob_2$ so that $\p_1(\x_S;\x'_{\Bar{S}}) = \p_2(\x_S;\x'_{\Bar{S}})$ and $\prob_1(\x_S;\x'_{\Bar{S}}) = \prob_2(\x_S;\x'_{\Bar{S}})$, then $\f_{\x,\p_1,\prob_1}(S) =  \f_{\x,\p_2,\prob_2}(S)$.
\end{myprops}

\begin{proof}
By (Set-DS), we know that $\f_{\x,\p,\prob}(S)$ depends only on $\p(\x_S;\cdot)$ and $\prob(\x_S;\cdot)$. We now want to show that for $\p_1, \p_2, \prob_1, \prob_2 $ such that $\p_1(\x_S;\x'_{\bar{S}}) = \p_2(\x_S;\x'_{\bar{S}})$ and $\prob_1(\x_S;\x'_{\Bar{S}}) = \prob_2(\x_S;\x'_{\Bar{S}})$, but $\p_1(\x_S;\x''_{\bar{S}}) \not = \p_2(\x_S;\x''_{\bar{S}})$ or $\prob_1(\x_S;\x''_{\bar{S}}) \not = \prob_2(\x_S;\x''_{\bar{S}})$, where $\x''$ is some arbitrary input not equal to $\x'$, $\f_{\x,\p_1,\prob_1}(S) =  \f_{\x,\p_2,\prob_2}(S)$. Thus, proposition \ref{prop:step2} is much stricter than (Set-DS).


Given any arbitrary $\p_1, \p_2, \prob_1, \prob_2$, we now construct $\p'_1$, $\p'_2$, $\prob'_1$, $\prob'_2$ by setting $f_1'(u):=f_1(u|\text{do } u_S=\x_S)$, $p_1'(u)=p_1(u|\text{do }u_S=\x_S)$, $f_2'(u):=f_2(u|\text{do }u_S=\x_S)$, $p_2'(u)=p_2(u|\text{do }u_S=\x_S)$. By (Set-DS), we know that $\f_{\x,\p_1,\prob_1}(S)  =  \f_{\x,\p'_1,\prob'_1}(S)$ and $\f_{\x,\p_2,\prob_2}(S)  =  \f_{\x,\p'_2,\prob'_2}(S)$ (*). We also clearly see that $\p_1', \p_2', \prob_1', \prob_2'$ are dummy in all features in set S. Thus, by (D-DS), we have $\f_{\x,\p'_1,\prob'_1}(S) = \f_{\x,\p'_1,\prob'_1}(\emptyset)$, $\f_{\x,\p'_2,\prob'_2}(S) = \f_{\x,\p'_2,\prob'_2}(\emptyset)$ (**). By (N-DS), we know $\f_{\x,\p'_1,\prob'_1}(\emptyset)$ depends only on $\p_1'(\x')$ and $\prob_1'(\x')$
which is $\p_1(\x_S;\x'_{\bar{S}})$ and $\prob_1(\x_S;\x'_{\bar{S}})$. Similarly, $\f_{\x,\p'_2,\prob'_2}(\emptyset)$ depends only on $\p_2(\x_S;\x'_{\bar{S}})$ and $\prob_2(\x_S;\x'_{\bar{S}})$. Therefore, if $\p_1, \p_2, \prob_1, \prob_2$ satisfies $\p_1(\x_S;\x'_{\Bar{S}}) = \p_2(\x_S;\x'_{\Bar{S}})$ and $\prob_1(\x_S;\x'_{\Bar{S}}) = \prob_2(\x_S;\x'_{\Bar{S}})$, it follows that $\f_{\x,\p'_1,\prob'_1}(\emptyset) = \f_{\x,\p'_2,\prob'_2}(\emptyset)$. Combining this with the properties (*) and (**) allows us to conclude $\f_{\x,\p_1,\prob_1}(S) =  \f_{\x,\p_2,\prob_2}(S)$, which completes our proof.

\end{proof}

\paragraph{Final Proof of Thm. \ref{thm:JBshap-uniqueness}}
With the proposition in hand, we are ready to prove Theorem \ref{thm:JBshap-uniqueness}.

The linearity property of  $\f$ (with respect to $\p, \prob$) implies $\f_{\x,k\p,\prob}(S) = k \f_{\x,\p,\prob}(S)$,  $\f_{\x,\p,k\prob}(S) = k \f_{\x,\p,\prob}(S)$ for any $k\in \R$. Now, by proposition \ref{prop:step2}, we can write $\f_{\x, \p,\prob}(S) = G(\p(\x_S;\x'_{\Bar{S}}),\prob(\x_S;\x'_{\Bar{S}}))$ for some functional $G$. 
Let us define $A_\p(\x,S) := \p(\x_S;\x'_{\Bar{S}})$ and $ B_\prob(\x,S) := \prob(\x_S;\x'_{\Bar{S}})$. 

The linearity property yields a key observation: given any arbitrary pair $(\x_0,S_0)$ and two functions $f_1, f_2$, we have 
\begin{align*}
    &G(A_{f_1}(\x_0,S_0)A_{f_2}(\x_0,S_0),B_p(\x_0,S_0)) \\
    &= A_{f_1}(\x_0,S_0) G( A_{f_2}(\x_0,S_0),B_p(\x_0,S_0)).
\end{align*}

%
The above holds for any choice of $f_1, f_2$ given a fixed pair $(\x_0, S_0)$. We may henceforth select $f_2 \equiv 1$, $\x = \x_0$, $S = S_0$, yielding 
\begin{align*}
&G(A_{f_1}(\x_0,S_0), B_p(\x_0,S_0)) \\
&= A_{f_1}(\x_0,S_0) G( 1,B_p(\x_0,S_0)).
\end{align*}
A similar reasoning gives for any choice $f,p$
\begin{align*}
    &G(A_f(\x_0,S_0), B_p(\x_0,S_0)) \\
    &= B_p(\x_0,S_0) G( A_f(\x_0,S_0),1),
\end{align*}
and consequently
\begin{align*}
    &G(A_f(\x_0,S_0), B_p(\x_0,S_0)) \\
    &= A_f(\x_0,S_0) B_p(\x_0,S_0) G(1,1)
\end{align*}
holds for any pair $(x_0, S_0)$. It then follows that 
\begin{align*}
    \f_{\x, \p,\prob}(S) &= G(A_f(\x,S), B_p(\x,S)) \\
    &= C  \p(\x_S;\x'_{\Bar{S}}) \prob(\x_S;\x'_{\Bar{S}}),
\end{align*}
where $C: = G(1,1)$. 
Finally, with the efficiency axiom, we see that only $C = 1$ satisfies the axiom. This concludes one direction of Theorem \ref{thm:JBshap-uniqueness}.

We remark that the uniqueness of the value function (up to a constant) does not rely on (Eff.), (Sym.), (Rob.) as they are mostly not used in the proof of the first part, and thus can be rewritten as a stronger theoretical result. However, we include them in the Theorem to make the theoretical statement more concise and emphasize that JBShap satisfies these desired properties.

Moving onwards, it remains to prove that  $\f_{\x,\p,\prob}(S)$ satisfies the (Lin.), (Sym.), (Dum.), (Null), (Eff.), (Set.), (Rob.) axioms if $\f_{\x,\p,\prob}(S) = \p(\x_S;\x'_{\Bar{S}})\prob(\x_S;\x'_{\Bar{S}})$. It is easy to see that the first six axioms hold. Robustness is shown by fixing any functions $f_1,f_2,p$ and noting 

\begingroup
\allowdisplaybreaks
\begin{align*}
    &(f_1(\x_S;\x'_{\Bar{S}})p(\x_S;\x'_{\Bar{S}}) - f_2(\x_S;\x'_{\Bar{S}})p(\x_S;\x'_{\Bar{S}})) \\
    &= (f_1(\x_S;\x'_{\Bar{S}}) - f_2(\x_S;\x'_{\Bar{S}}))p(\x_S;\x'_{\Bar{S}})\\
    &\leq \max_{\x} |(f_1(\x)-f_2(\x))|p(\x) \\
    &\leq \frac{1}{T}\max_{\x} |f_1(\x)-f_2(\x))|p(\x).
\end{align*}
\endgroup
As expected, strong $T-$robustness is satisfied when $T \leq 1$.





\subsection{Proof of Theorem \ref{thm:thm_transfer_1}}

Theorem \ref{thm:thm_transfer_1} is proved through the following six straightforward propositions, where the proofs simply follow by the definition.

\begin{myprops}
If two given input function and set function pair $(\p, \prob, \f_{\x,\p,\prob}(\cdot))$ satisfies (L-DS), and set function and explanation pair $(\f_{\x,\p,\prob}(\cdot), \ex(\f_{\x,\p,\prob}(\cdot)) )$ satisfies (L-SE), then $(\p, \prob, \ex(\f_{\x,\p,\prob}(\cdot)))$ satisfy (L-DE).
\end{myprops}

\begin{myprops}
If two given input function and set function pair $(\p, \prob, \f_{\x,\p,\prob}(\cdot))$ satisfies (S-DS), and set function and explanation pair $(\f_{\x,\p,\prob}(\cdot), \ex(\f_{\x,\p,\prob}(\cdot)) )$ satisfies (S-SE), then $(\p, \prob, \ex(\f_{\x,\p,\prob}(\cdot)))$ satisfy (S-DE).
\end{myprops}

\begin{myprops}
If two given input function and set function pair $(\p, \prob, \f_{\x,\p,\prob}(\cdot))$ satisfies (D-DS), and set function and explanation pair $(\f_{\x,\p,\prob}(\cdot), \ex(\f_{\x,\p,\prob}(\cdot)) )$ satisfies (D-SE), then $(\p, \prob, \ex(\f_{\x,\p,\prob}(\cdot)))$ satisfy (D-DE).
\end{myprops}

\begin{myprops}
If two given input function and set function pair $(\p, \prob, \f_{\x,\p,\prob}(\cdot))$ satisfies (E-DS), and set function and explanation pair $(\f_{\x,\p,\prob}(\cdot), \ex(\f_{\x,\p,\prob}(\cdot)) )$ satisfies (E-SE), then $(\p, \prob, \ex(\f_{\x,\p,\prob}(\cdot)))$ satisfy (E-DE).
\end{myprops}

By combining all propositions above, we conclude Theorem \ref{thm:thm_transfer_1}.

\end{document}